\pdfoutput=1
\documentclass[12pt]{article}
\usepackage[margin=1in]{geometry}
\usepackage{amsmath,amssymb}
\usepackage[super,sort&compress]{natbib}

\usepackage{booktabs}
\usepackage{graphicx}
\usepackage{hyperref}
\usepackage{authblk}
\usepackage{threeparttable}
\usepackage{subcaption}
\usepackage{xurl}
\usepackage{textcomp}
\usepackage[inline]{enumitem}
\graphicspath{{./}}

\title{%
  A Mixed-Stiffness Anthropomimetic Fingertip Broadens the Operating Range for Coin Grasping\\[6pt]
}

\author[1]{Kaigen Go}
\author[2]{Yinlai Jiang}
\author[1,2]{Hiroshi Yokoi}
\author[1,2]{Shunta Togo\thanks{Corresponding author: Shunta Togo, Graduate School of Informatics and Engineering, The University of Electro-Communications, 1-5-1 Chofugaoka, Chofu, Tokyo 182-8585, Japan. Email: \href{mailto:s.togo@uec.ac.jp}{s.togo@uec.ac.jp}}}
\affil[1]{Graduate School of Informatics and Engineering, The University of Electro-Communications, Tokyo, Japan}
\affil[2]{Center for Neuroscience and Biomedical Engineering, The University of Electro-Communications, Tokyo, Japan}

\date{}

\begin{document}
\maketitle

\begin{abstract}
Robotic grasping of thin, flat objects such as coins on hard surfaces remains challenging because conventional methods require reorienting the object, accessing its underside, or adding a dedicated nail mechanism. We previously showed that a rigid nail arrests soft-pad deformation and thereby forms a geometric constraint that improves precision grasping. Here we asked whether an additional constraint-forming boundary, created within the pad by material choice rather than by anatomy, could extend the conditions under which that constraint holds.
We fabricated anthropomimetic fingertips with Shore E10 silicone at the center and Shore A60 at the sides, and compared them with uniformly soft E10 fingertips. An automated apparatus performed an oblique rotational tip pinch in which the pad engaged the coin's lateral surface, lifting it from flush contact with no gap beneath it. Over variations in horizontal approach distances, vertical finger displacements, and index-finger rotation, the mixed-stiffness pair maintained high success rates across more tested settings than the uniform pair during both geometric-constraint formation and the transition to a stable grasp. The nail-free pair failed in all 36 conditions of Experiment~1-1. However, the uniform pair performed better when coin position along the finger axis was varied, a condition-dependent trade-off. After tuning for coin size, both fingertip types grasped all six Japanese denominations.
These results suggest that the operating range for thin-object grasping depends not only on pad softness but also on where stiffness is placed within a nail-supported pad, making boundary placement a candidate fingertip design variable.
\end{abstract}


\section{Introduction}\label{sec:intro}

\begin{figure}[tb]
    \centering
    \includegraphics[width=0.9\linewidth]{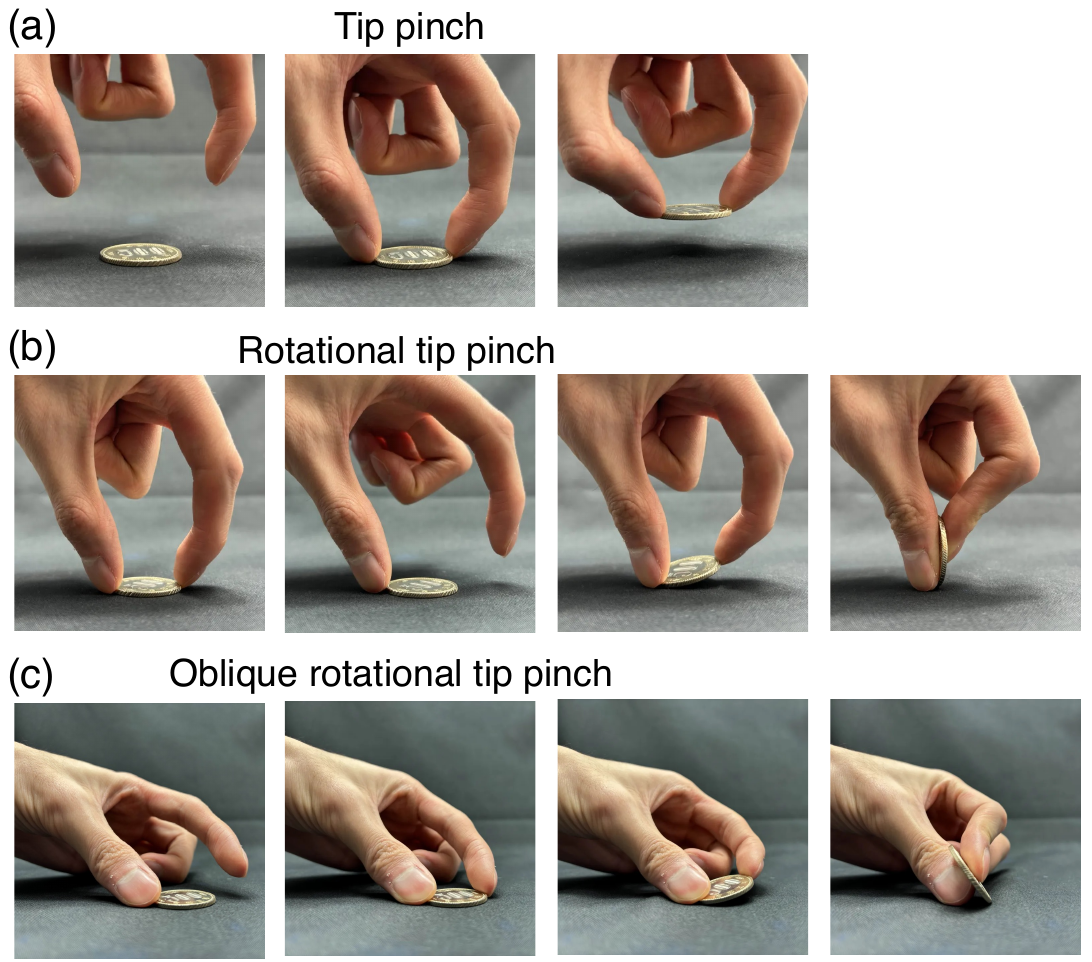}
    \caption{Three pinch strategies for picking up a coin. Each sequence shows the motion progression from left to right. (a)~Tip pinch. (b)~Rotational tip pinch. (c)~Oblique rotational tip pinch.}
    \label{fig:posture}
\end{figure}

Robotic grasping of thin, flat objects such as coins from a hard surface remains difficult in industrial, assistive, and prosthetic applications.\cite{T_Yoshimi_12, Z_Tong_20, T_He_21} Many robotic strategies resemble the tip or rotational tip pinches in Fig.~\ref{fig:posture}a,b and require either access to the object's underside, often through a nail or scooping structure, or prior reorientation of the object. Reduced fingernail length impairs fine-object manipulation during the human tip pinch,\cite{CWS_Jansen_00, R_Shirato_17} suggesting that the strategies in Fig.~\ref{fig:posture}a,b partly rely on the nail engaging the object's lower edge. Humans, however, need not work from beneath: a coin can be picked up by tilting a finger, pressing the pad against the coin's lateral edge, and rotating the coin off the surface (Fig.~\ref{fig:posture}c). In this posture, the tilted pad envelops the coin's lateral surface, and the nail is never inserted beneath it. This strategy has received little attention in robotic grasping research.

The human fingertip comprises bone, nail, skin, and subcutaneous tissue, and artificial fingers reproducing combinations of these elements can improve grasping performance.\cite{KB_Shimoga_96, K_Hosoda_06, M_Controzzi_14, K_Or_16} However, most prior work has focused on individual anatomical components or overall fingertip compliance rather than regional stiffness distribution within the pad. Anatomical studies indicate that the lateral sides of the human fingertip are stiffer than the central pad, partly because of the proper digital ligaments.\cite{A_Perez-Gonzalez_13, A_Gupta_21} In our previous anthropomimetic fingertip, a rigid nail arrested soft-pad deformation and formed a geometric constraint, increasing grasp force by up to 3.10 times in a block-grasping task.\cite{A_Kumagai_23} In that case the constraint arose at an anatomical boundary between the rigid nail and the soft pad, and acted dorsally across the entire fingertip. Here we ask whether an additional constraint-forming boundary can be created within the pad surface itself, not by anatomy but by material choice. If so, stiff lateral regions flanking a soft central pad should confine deformation near the coin edge and sustain constraint under settings where a uniformly soft pad fails. The location of such boundaries would then become a design variable, whereas the nail is fixed by the anatomy the fingertip reproduces. We therefore focus not on the presence or absence of the nail itself, but on how the arrangement of lateral stiffness modifies constraint formation in a nail-supported fingertip.

In this study, we tested whether the nail--pad constraint principle extends to a localized center--lateral boundary. We fabricated a mixed-stiffness anthropomimetic fingertip with Shore~E10 silicone centrally and Shore~A60 silicone laterally, and compared it with a uniformly soft E10 fingertip using an automated oblique rotational tip pinch. Performance was evaluated over variations in horizontal approach distance, vertical displacement, index-finger rotation, and coin position. Rather than examining only peak success under tuned settings, we quantified the successful operating range across the discrete tested conditions. We also separated geometric-constraint formation from the transition to a stable grasp because initial constraint does not guarantee successful completion. This work makes three contributions:
\begin{itemize}[nosep, topsep=4pt, leftmargin=1.5em]
    \item We introduce passive lateral stiffness patterning as a fingertip design axis.
    \item We quantify the stage-specific effect of that patterning on the successful operating range.
    \item We identify a parameter variation for which the uniform fingertip performs better, revealing a condition-dependent trade-off.
\end{itemize}

\section{Related Work}\label{sec:related}

\subsection{Grasping Methods for Thin, Flat Objects}

Robotic methods for picking up thin, flat objects from a hard surface follow the tip and rotational tip pinches of Fig.~\ref{fig:posture}a,b and differ mainly in how they obtain purchase at the object's edge. Most insert a nail or slender digit into the object--surface interface: hooking a soft nail at the object's side after sliding it,\cite{T_Yoshimi_12} tilting the object against a fixture and tucking a finger into the resulting gap,\cite{Z_Tong_20} scooping with a variable-length digit,\cite{T_He_21} or deploying a retractable\cite{S_Jain_20} or pulp-backed rigid\cite{DH_Kang_26} fingernail; when the object is rotated, the rotation opens or exploits that interface rather than replacing it. Two studies dispense with insertion. Matsuno et al.\cite{T_Matsuno_05} generated lift from soft-fingertip deformation alone, though the object was rotated rather than brought clear of the surface. Closest to the present work, Odhner et al.\cite{L_Odhner_13} pinned a coin at one edge with the thumb and swept an opposing underactuated finger along the table to flip it into a pinch grasp; small ridges on the pads were required to catch the coin's edge at a favorable contact angle, and success degraded below a coin thickness of about 2~mm because those ridges could not be fabricated small enough. The present oblique rotational tip pinch also avoids insertion and prior reorientation. It is instead designed to exploit a stiffness boundary within the smooth pad, and we evaluate constraint formation and the subsequent transition as separate stages.

\subsection{Fingertip Stiffness Design}

Fingertip stiffness has been designed along four axes. Through-thickness or radial layering treats depth as the design variable, as in a rigid core combined with polymer layers of differing stiffness and a hard nail,\cite{M_Controzzi_14} a two-layer elastic fingertip,\cite{Y_Obata_20} and our own nail--soft-tissue construction.\cite{A_Kumagai_23} Localized constraint by anatomical structure places rigid elements at nails or skeletal features, as in asymmetric distal-phalanx geometry\cite{A_Kumagai_21} and studies of how nail presence affects grasp stability\cite{K_Or_16}; the positions of these elements are fixed by the anatomy being reproduced. Distributed compliance acts at the scale of the finger or hand rather than within the pad, spanning skin, finger, and wrist,\cite{K_Junge_25} internal gradient lattices,\cite{SJ_Schouten_25} rigid--flexible--soft hierarchies,\cite{B_Lyu_25} and skeleton--ligament--tendon hybrids.\cite{N_Zhang_25} Active variable stiffness modulates stiffness over time using jamming, shape-memory alloys, low-melting-point alloys, or magnetorheological fluids,\cite{Y_Shan_24, Z_Li_24, H_Li_24} a different design philosophy from a passive, time-invariant spatial arrangement. Lateral stiffness patterning across the pad surface, particularly a soft-center/stiff-side distribution, has received limited attention.

The closest precedent is Bullock et al.,\cite{IM_Bullock_15} who gave the finger a ridged rigid core with intervening air gaps beneath a uniform silicone skin; the ridged designs outperformed a conventional solid core in manipulation stability. There the pattern arises from internal core geometry beneath a single-material skin and repeats around the finger, and it was evaluated for in-hand manipulation. The present fingertip instead patterns the pad material itself, in a two-material soft-center/stiff-side arrangement drawn from fingertip anatomy, and is assessed by its successful operating range for thin-object pickup. To our knowledge, no study has compared such an arrangement with a uniformly soft fingertip on that criterion.

\section{Methods}\label{sec:methods}

\subsection{Task Definition and Evaluation Strategy}\label{sec:task}

The task was to lift a coin from a hard, flat surface and transition to a stable two-finger grasp using the oblique rotational tip pinch (Fig.~\ref{fig:posture}c). The coin lay flush against the plate at the start of every trial, so lifting was initiated without any gap beneath it. Following our previous work,\cite{A_Kumagai_23} we define \textit{geometric constraint} as the effective restriction of coin motion when soft-pad deformation is arrested by the nail or stiff lateral regions; it is distinct from rigid-body form closure. The successful operating range denotes the distribution of high-success settings among the discrete tested conditions, not a continuous geometric area.

\subsection{Artificial Fingers and Stiffness Distribution}

\begin{figure}[ptb]
    \centering
    \includegraphics[width=0.6\linewidth]{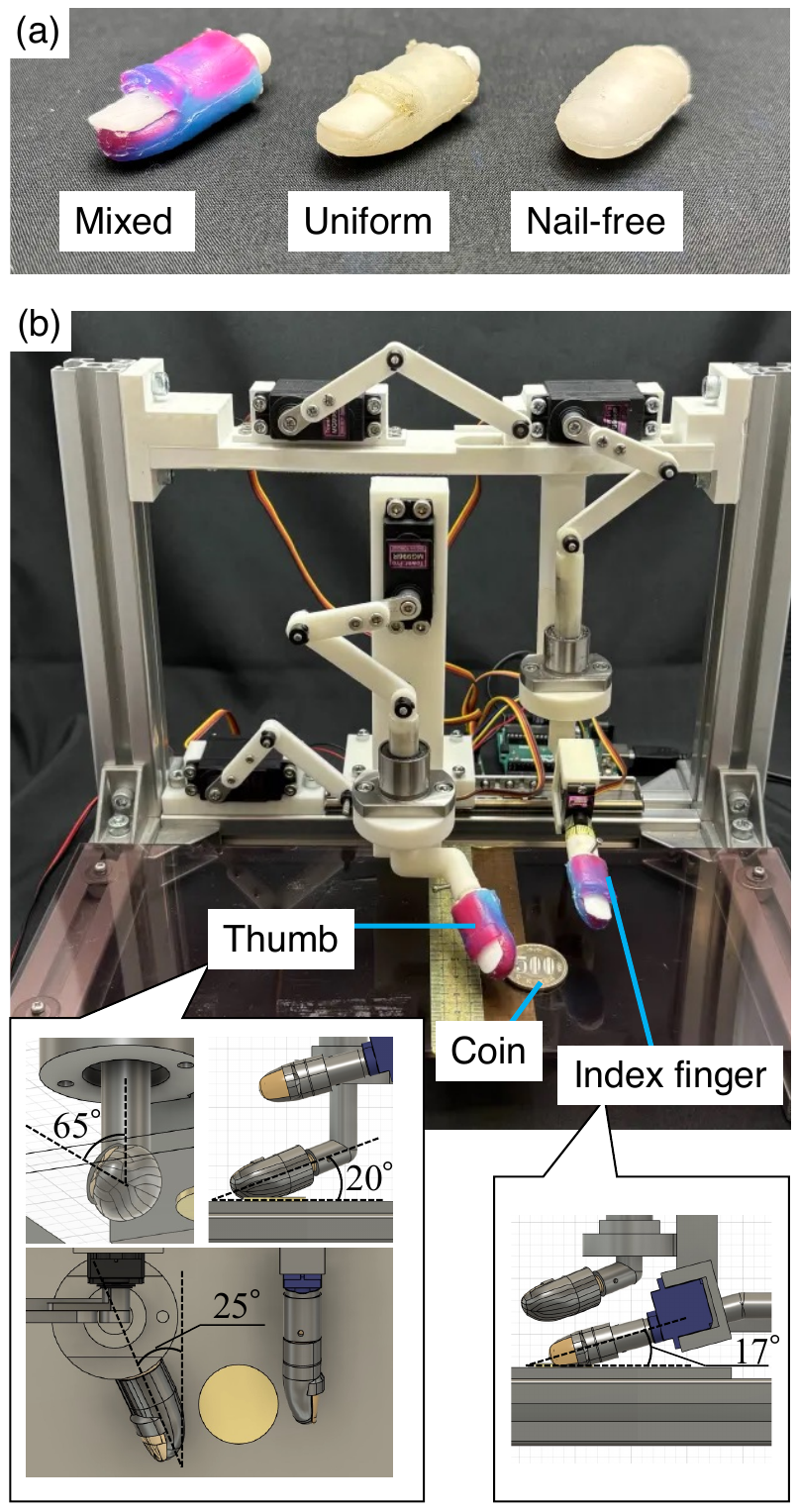}
    \caption{Overview of the experimental setup. (a)~Three types of artificial fingertips compared: the mixed fingertip with a soft center and stiff sides, the uniform fingertip with homogeneous E10 skin, and the nail-free fingertip. In the mixed fingertip, blue regions indicate E10 skin, magenta regions indicate A60 skin, and white areas denote the nail. (b)~Automated grasping apparatus and the arrangement angles of the thumb and index finger.}
    \label{fig:apparatus}
\end{figure}

All fingers reproduced bone, skin, and subcutaneous tissue using the structure of our previous designs,\cite{A_Kumagai_21, A_Kumagai_23} and all but the nail-free fingertip also carried a nail. Table~\ref{tab:conditions} and Fig.~\ref{fig:apparatus}a summarize the three fingertip types. Bone and nail were stereolithographically printed (Form~3, Formlabs, Inc., USA; Standard White Resin~V4), and Soma Foama~15 (Smooth-On, Inc., USA) formed the subcutaneous tissue. Skin thickness was 1.5~mm, based on human fingertip cross-sections\cite{T_Maeno_98, H_Fruhstorfer_00} and on fabrication reproducibility. Geometry, internal tissue, dimensions, and surface coating (Rita-Surf TSM-2, Rita Fine, Japan) were identical across conditions.

The uniform skin was Shore~E10 silicone (Toughsilon Gel TSG-E10, Tanac Co. Ltd., Japan; Supplementary Fig.~\ref{fig:supp_uniform_fabrication}), the softest skin material in our previous work, where the nail-mediated geometric-constraint effect was most pronounced.\cite{A_Kumagai_23} The mixed skin combined E10 centrally with Shore~A60 silicone (TSE3466, Tanac Co. Ltd., Japan) laterally and was molded through separate injection channels (Supplementary Fig.~\ref{fig:supp_mixed_fabrication}). In prior measurements using the same internal tissue, the apparent Young's moduli were $77.9 \pm 12.0$~kPa for a fingertip with E10 skin and $1302.6 \pm 138.1$~kPa for one with A60 skin, a ratio of approximately 17.\cite{A_Kumagai_23} The present proof of concept tests the qualitative soft-center/stiff-side arrangement rather than an anatomically optimized ratio.

\begin{table}[tb]
\centering
\caption{Experimental conditions.}\label{tab:conditions}
\begin{tabular}{llll}
\toprule
Label & Skin material & Boundaries present & Usage \\
\midrule
Mixed     & E10 + A60 & Nail--pad, lateral & Primary condition \\
Uniform   & E10       & Nail--pad          & Control condition \\
Nail-free & E10       & Neither of the above & Experiment~1-1 only \\
\bottomrule
\end{tabular}
\end{table}

\subsection{Automated Grasping Apparatus}

To enable reproducible comparison of the artificial fingers, we built an automated apparatus that performs the oblique rotational tip pinch (Fig.~\ref{fig:apparatus}b). Four MG996R servomotors (Tower Pro Pte Ltd, Singapore) drove four linear-link mechanisms for horizontal and vertical motion of the two fingers, and an MG90D servomotor (Tower Pro Pte Ltd, Singapore) rotated the index finger. An Arduino UNO R3 (Arduino, Italy) and PCA9685 driver (NXP Semiconductors, The Netherlands) controlled all actuators; the control program is provided as Code~S2. The fingers were arranged to reproduce the human grasp posture; the arrangement angles were $17^\circ$ (index finger), $20^\circ$ (thumb), $65^\circ$ (thumb axis), and $25^\circ$ (coin-side surface), as defined in the figure.

\begin{figure}[!t]
    \centering
    \includegraphics[width=0.8\linewidth]{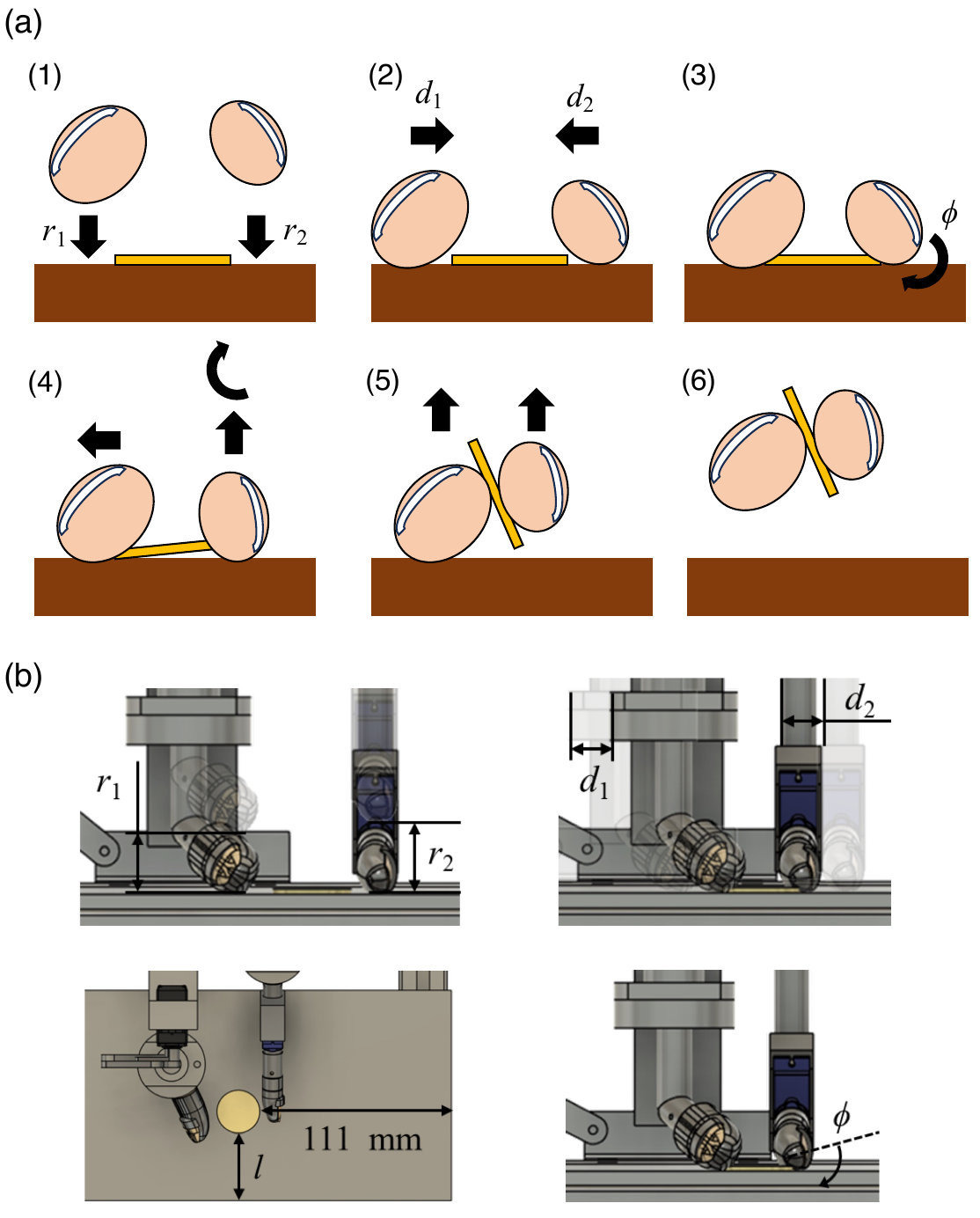}
    \caption{Overview of the experimental procedure. (a)~Six-stage motion sequence of the oblique rotational tip pinch. Stages~1--3 correspond to geometric-constraint formation; stages~4--6 correspond to transition to a stable grasp. (b)~Definitions of vertical finger displacements $r_1$, $r_2$, horizontal approach distances $d_1$, $d_2$, index-finger rotation angle $\phi$, and coin position $l$. Distance units are mm; angle units are degrees.}
    \label{fig:sequence}
\end{figure}

The six stages were as follows: (1) vertical displacement of both fingers toward the flat surface from their initial positions (commanded displacements $r_1,r_2$); (2) horizontal approach toward the coin ($d_1,d_2$); (3) index rotation $\phi$ to lift the coin; (4) thumb retraction by 8~mm and index rise by 13~mm; (5) thumb and index rises of 14 and 12~mm; and (6) holding the final grasp posture (Fig.~\ref{fig:sequence}). Stages~1--3 formed the geometric constraint, and stages~4--6 transitioned to a stable grasp; the stage~4--6 motions were determined by trial and error, and the values given here are those used in Experiments~1 and~2. Coin position $l$ was the distance from the lower edge of the PVC plate to the lower edge of the coin (Fig.~\ref{fig:sequence}b). The horizontal distance from the plate's right edge to the coin's right edge was fixed at 111~mm. The CAD and 3D-printable part files for the apparatus and the fingertip molds are available in the GitHub repository (see Data Availability).

\subsection{Experimental Procedure and Test Conditions}

A coin was positioned using a graduated reference, and one automated trial was executed. The coin was then returned to its initial position, and both the coin and the fingertip surfaces were inspected before the next trial. Ten trials were conducted per condition and recorded as binary outcomes; partial completion was classified as a failure.

Experiments~1 and~2 used a current Japanese 500-yen coin (diameter 26.5~mm, thickness 1.8~mm). Experiment~1 explored geometric-constraint formation; success required the coin to remain lifted for at least 2~s after stage~3. Table~\ref{tab:exp1} lists the explored and fixed parameters; the measured $d_1$--$d_2$ grid was slightly nonuniform because of servo resolution. The nail-free pair was tested only in Experiment~1-1 and omitted thereafter because it failed in all conditions. Fixed settings for Experiments~1-2--1-4 were selected from high-success conditions in Experiment~1-1.

\begin{table}[tb]
\centering
\caption{Conditions for Experiment~1.}\label{tab:exp1}
\footnotesize
\setlength{\tabcolsep}{4pt}
\begin{threeparttable}
\begin{tabular}{cll}
\toprule
Exp. & Explored parameters & Fixed parameters \\
\midrule
1-1 & $d_1=\{0,4,8,12,15,19\}$\,mm, $d_2=\{0,4,9,13,16,20\}$\,mm & $r_1=16$, $r_2=33$, $\phi=40$, $l=32$ \\
1-2 & $r_1=\{10,15,20\}$\,mm, $r_2=\{13,18,23\}$\,mm & $d_1=8$, $d_2=13$, $\phi=40$, $l=32$ \\
1-3 & $\phi=\{25,30,35,40\}^\circ$ & $d_1=8$, $d_2=13$, $r_1=15$, $r_2=23$, $l=32$ \\
1-4 & $l=\{28,30,32,34,36,38,40\}$\,mm & $d_1=8$, $d_2=13$, $r_1=15$, $r_2=23$, $\phi=40$\\
\bottomrule
\end{tabular}
\begin{tablenotes}
\footnotesize
\item Ten trials were conducted for each condition in all experiments.
\end{tablenotes}
\end{threeparttable}
\end{table}

Experiment~2 evaluated the transition through stages~4--6 for settings with at least 80\% geometric-constraint success in Experiment~1-1. Success required the coin to complete the sequence without being dropped, re-contacting the surface, or slipping from between the fingers, and then to be held stably for at least 5~s. We tested 20 mixed and 17 uniform conditions; inferential comparison used the 14 common conditions, whereas reductions in high-success conditions within each fingertip type used all settings tested for that type.

Experiment~3 applied the same stable-grasp criterion to all six Japanese coin denominations (diameters 20.0--26.5~mm, thicknesses 1.5--1.8~mm). Only the initial separation between the fingers and the stage-4 index rise were adjusted for coin diameter; $d_1=8$~mm, $d_2=13$~mm, $r_1=15$~mm, $r_2=23$~mm, $l=32$~mm, and $\phi=40^\circ$ were fixed.

\subsection{Analysis Methods}

Condition-wise success rates were visualized as heat maps. Robustness to parameter variation was summarized by the integrated success score,\cite{P_Mannam_24}
\begin{equation}
S=\sum_{i=1}^{N}p_i,
\label{eq:isr}
\end{equation}
where $p_i$ is the success rate at condition $i$ and $N$ the number of tested conditions. $S/N$ is the mean success rate, and $\Delta S=S_{\mathrm{mixed}}-S_{\mathrm{uniform}}$ was the primary effect measure.

A parametric bootstrap\cite{RJ_Tibshirani_93} generated success counts from $\mathrm{Binomial}(10,\hat p_i)$ for 10{,}000 samples to estimate a 95\% confidence interval for $\Delta S$. Because plug-in probabilities of 0 or 1 may underestimate uncertainty, the intervals were treated as approximate indicators for the tested specimens and conditions.

For Experiments~1-1 and~2, a Cochran--Mantel--Haenszel test compared success and failure between fingertip types, treating each parameter condition as a stratum containing a $2\times2$ table of outcome by fingertip type.\cite{N_Mantel_59} The Breslow--Day test assessed odds-ratio homogeneity; when homogeneity was rejected, the Mantel--Haenszel common odds ratio was reported only descriptively. Strata in which both fingertip types yielded identical all-success or all-failure outcomes carried no information about the odds ratio and were excluded from the Breslow--Day statistic, so its degrees of freedom are smaller than the total number of tested conditions minus one. The significance level was $\alpha=0.05$. Experiments~1-2--1-4 were evaluated with $\Delta S$ and bootstrap intervals because they had fewer strata.

To check that the difference in operating range did not depend on the chosen success-rate threshold, a threshold sweep counted conditions meeting $\tau=0.1,0.2,\ldots,1.0$, plus conditions with any success\cite{P_Mannam_24} (Supplementary Fig.~S3). Sensitivity to nonuniform $d_1$--$d_2$ spacing was assessed with midpoint-derived normalized cell-area weights for Experiments~1-1 and~2 (Supplementary Table~S1). The analysis code and the condition-wise trial outcomes are provided as Code~S1.

\section{Results}\label{sec:results}

Trials with the automated apparatus confirmed that coins could be lifted from the flat surface and transitioned to a stable grasp via the oblique rotational tip pinch, as shown in Fig.~\ref{fig:grasping} and Video~S1.
Quantitative results for each experimental parameter are presented below.

\begin{figure}[!htbp]
    \centering
    \includegraphics[width=0.5\linewidth]{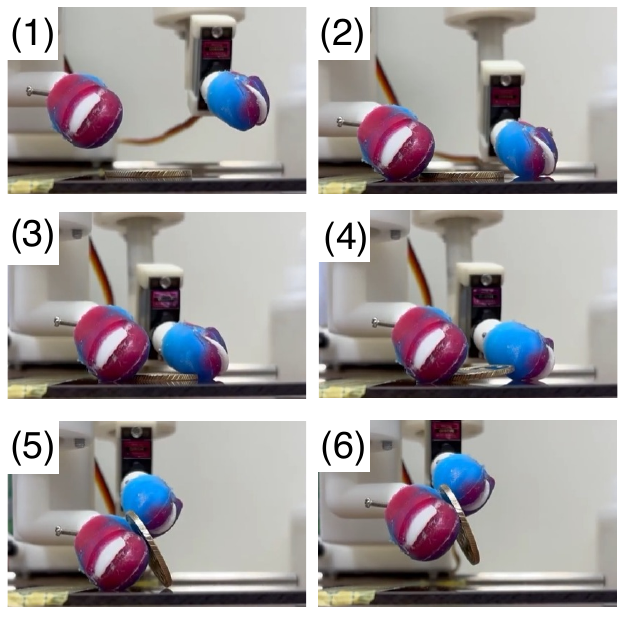}
    \caption{Grasping sequence of a Japanese 500-yen coin using the mixed artificial-finger pair. Panels~(1)--(6) correspond to the motion stages shown in Fig.~\ref{fig:sequence}a.}
    \label{fig:grasping}
\end{figure}

\subsection{Experiment~1-1: Horizontal Approach Distances $d_1$, $d_2$ and Geometric-Constraint Formation}

\begin{figure}[!htbp]
    \centering
    \includegraphics[width=0.8\linewidth]{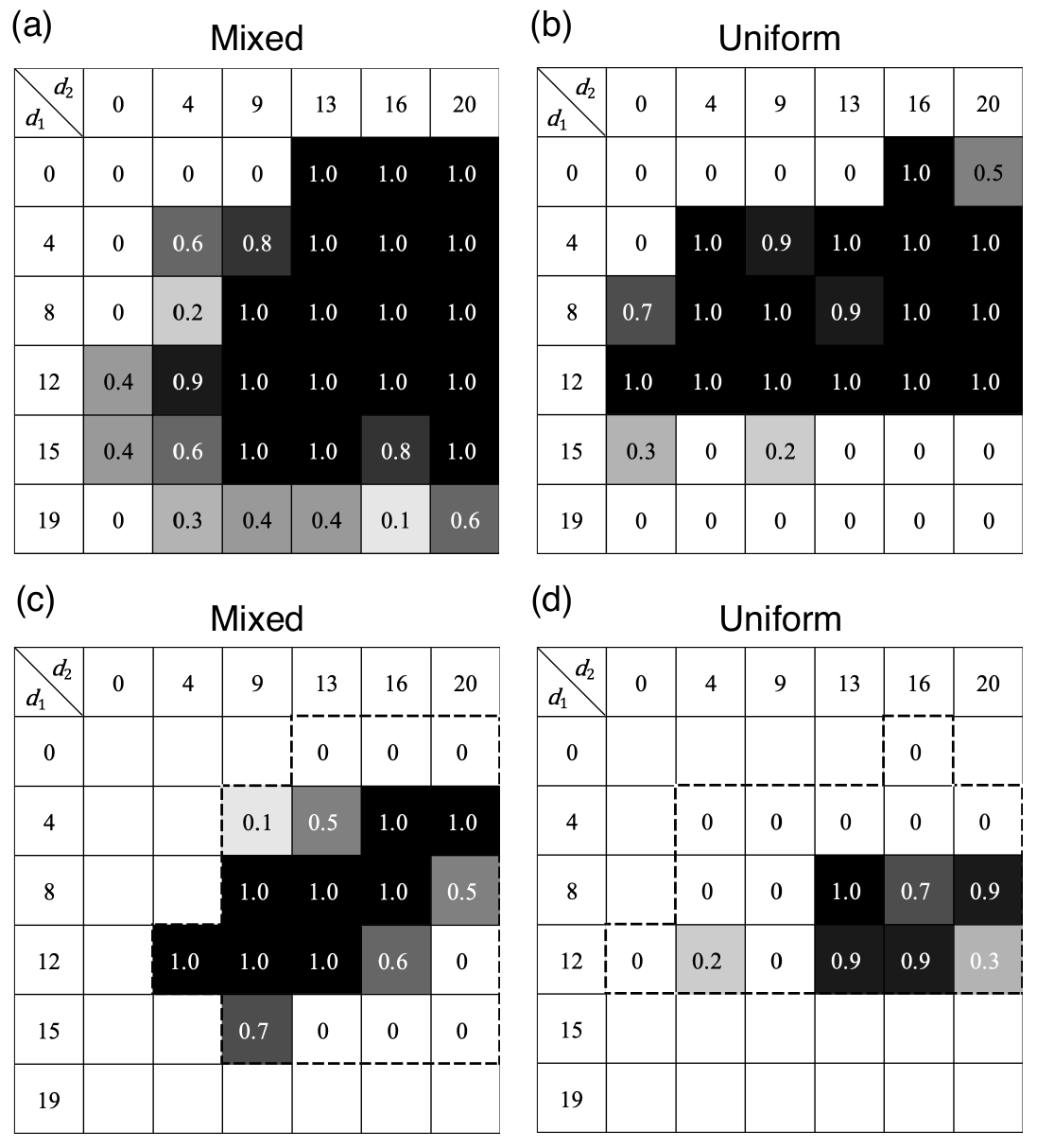}
    \caption{Results of Experiments~1-1 and~2. (a,~b)~Geometric-constraint formation success rates for the mixed and uniform pairs. The number in each cell indicates the proportion of successes out of 10~trials; shading from white to black corresponds to success rates from 0 to~1. (c,~d)~Success rates for transition to a stable grasp for the mixed and uniform pairs. Blank cells indicate conditions for which this test was not conducted. Dashed lines demarcate conditions where the geometric-constraint formation success rate was 80\% or higher in Experiment~1-1. Statistical comparison between fingertip conditions used the 14~conditions with values in both~(c) and~(d). Units of $d_1$ and $d_2$ are mm.}
    \label{fig:exp_1-1_2}
\end{figure}

Figures~\ref{fig:exp_1-1_2}a and~b show the geometric-constraint formation success rates for the mixed and uniform pairs on the $d_1$--$d_2$ parameter plane ($6 \times 6 = 36$~conditions).
The nail-free pair failed to form a geometric constraint in all 36~conditions.

Aggregate success across the tested $d_1$--$d_2$ grid was higher for the mixed pair: the integrated success score was 23.5 (mean: 0.653) for the mixed pair and 18.5 (mean: 0.514) for the uniform pair; the bootstrap 95\% confidence interval for the difference $\Delta S = 5.0$ was $[3.8,\; 6.2]$.
In the threshold-sweep analysis, the mixed pair showed equal or greater numbers of successful conditions at every threshold level from $\tau = 0.1$ to $\tau = 1.0$ (e.g., number of conditions with success rate $> 0$: mixed pair $= 30$, uniform pair $= 21$; number of conditions with success rate $\geq 0.8$: mixed pair $= 20$, uniform pair $= 17$).
Across all examined thresholds, the mixed pair therefore exhibited a broader successful operating range in the $d_1$--$d_2$ parameter space.
The CMH test indicated that the success probability of the mixed pair was systematically and significantly higher than that of the uniform pair ($\chi^2(1) = 31.5$, $p < 0.001$).
However, the Breslow--Day test rejected the homogeneity of odds ratios across strata ($\chi^2(19) = 145.6$, $p < 0.001$), and the common odds ratio (MH estimator $= 2.69$, 95\% CI $[1.38,\; 5.25]$) was interpreted as a supplementary indicator because it was influenced by strata at the boundary of the parameter space.

Weighting the conditions by grid-cell area to account for the nonuniform $d_1$--$d_2$ spacing left the direction of the difference unchanged (mean success rate 0.632 vs.\ 0.510; Supplementary Table~S1).

\subsection{Experiment~1-2: Vertical Finger Displacements $r_1$, $r_2$ and Geometric-Constraint Formation}

\begin{figure}[!t]
    \centering
    \includegraphics[width=0.6\linewidth]{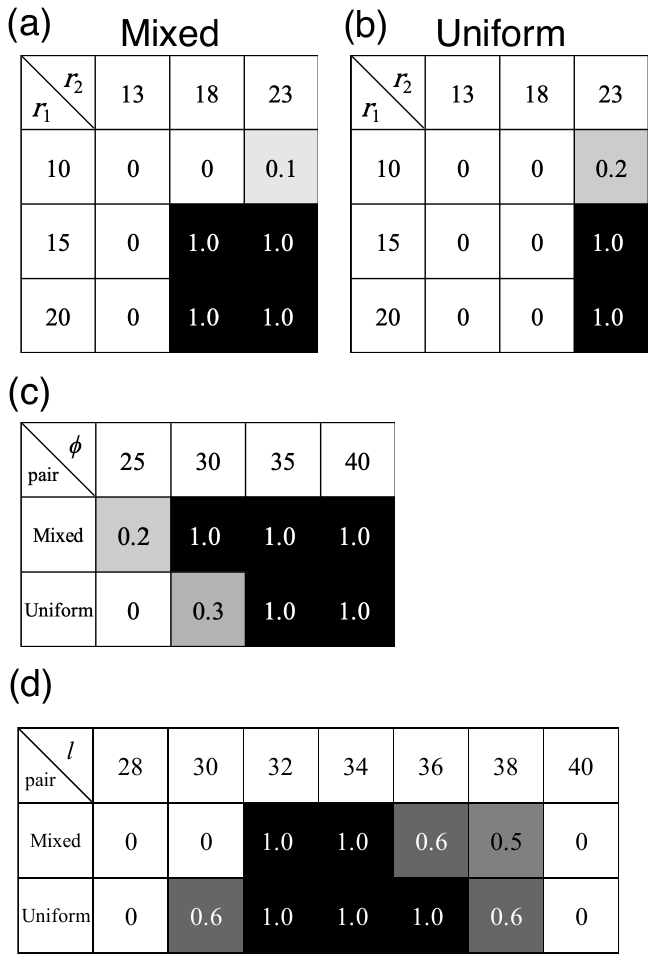}
    \caption{Geometric-constraint formation success rates in Experiments~1-2 through~1-4. (a,~b)~Results for vertical finger displacements $r_1$, $r_2$ for the mixed and uniform pairs. (c)~Results for index-finger rotation angle $\phi$. (d)~Results for coin position $l$. The number in each cell indicates the proportion of successes out of 10~trials; shading from white to black corresponds to success rates from 0 to~1. Units of $r_1$, $r_2$, $l$ are mm; units of $\phi$ are degrees.}
    \label{fig:exp_1-2_1-4}
\end{figure}

As shown in Figs.~\ref{fig:exp_1-2_1-4}a and~b, the mixed pair formed a geometric constraint at least once in 5~of 9~conditions, compared with 3~conditions for the uniform pair.
The integrated success score was 4.1 (mean: 0.456) for the mixed pair and 2.2 (mean: 0.244) for the uniform pair; the bootstrap 95\% confidence interval for the difference was $[1.6,\; 2.2]$.
The minimum $r_2$ required for geometric-constraint formation was larger for the uniform pair than for the mixed pair, indicating that the mixed pair tolerated a broader range of vertical finger displacements.

\subsection{Experiment~1-3: Index-Finger Rotation $\phi$ and Geometric-Constraint Formation}

As shown in Fig.~\ref{fig:exp_1-2_1-4}c, the mixed pair achieved a success rate of 0.2 at $\phi = 25^\circ$ and reached a success rate of 1.0 at $\phi = 30^\circ$.
The uniform pair, by contrast, showed a success rate of 0 at $\phi = 25^\circ$ and only 0.3 at $\phi = 30^\circ$, not reaching 1.0 until $\phi = 35^\circ$.
The integrated success score was 3.2 (mean: 0.800) for the mixed pair and 2.3 (mean: 0.575) for the uniform pair; the bootstrap 95\% confidence interval for the difference was $[0.5,\; 1.3]$.
These results indicate that the mixed pair could form a geometric constraint at a smaller rotation angle.

\subsection{Experiment~1-4: Coin Position $l$ and Geometric-Constraint Formation}

As shown in Fig.~\ref{fig:exp_1-2_1-4}d, the uniform pair exhibited higher success rates than the mixed pair when coin position $l$ was varied.
The integrated success score was 4.2 (mean: 0.600) for the uniform pair and 3.1 (mean: 0.443) for the mixed pair; the bootstrap 95\% confidence interval for the difference was $[-1.7,\; -0.5]$ (mixed pair $-$ uniform pair).
The threshold-sweep analysis also showed the uniform pair achieving equal or greater numbers of successful conditions at all thresholds.
Both fingertip pairs reached a success rate of 1.0 at $l = 32$--$34$\,mm and declined toward the extremes, but the uniform pair maintained high success rates over a broader range.

\subsection{Experiment~2: Transition to a Stable Grasp}

In Experiment~1-1, 20 conditions for the mixed pair and 17 for the uniform pair met the 80\% criterion and were tested for transition to a stable grasp.
Direct comparison using the 14~conditions tested with both fingertip pairs showed that the mixed pair achieved higher success rates for transition to a stable grasp (Figs.~\ref{fig:exp_1-1_2}c and~d).

The integrated success score was 9.7 (mean: 0.693) for the mixed pair and 4.9 (mean: 0.350) for the uniform pair; the bootstrap 95\% confidence interval for the difference $\Delta S = 4.8$ was $[4.0,\; 5.6]$.
The threshold-sweep analysis also showed the mixed pair achieving equal or greater numbers of successful conditions at all thresholds (e.g., number of conditions with success rate $= 1.0$: mixed pair $= 8$, uniform pair $= 1$).
The CMH test confirmed a significant difference in success probability ($\chi^2(1) = 48.2$, $p < 0.001$).
Because the Breslow--Day test rejected the homogeneity of odds ratios across strata ($\chi^2(11) = 84.9$, $p < 0.001$), the common odds ratio (MH estimator $= 5.32$, 95\% CI $[2.21,\; 12.80]$) was interpreted as a supplementary indicator.

Grid-area weighting again left the direction of the difference unchanged (mean success rate 0.692 vs.\ 0.325; Supplementary Table~S1).
The number of conditions with a success rate $\geq 80\%$ decreased by 60.0\% ($20 \to 8$ conditions) for the mixed pair and by 76.5\% ($17 \to 4$ conditions) for the uniform pair from geometric-constraint formation to the transition to a stable grasp.

\subsection{Experiment~3: Application to Coins of Different Sizes}

In this experiment, both the mixed and uniform pairs achieved a success rate of 1.0 for all six coin denominations.
This suggests that the individually tuned parameters were favorable enough for both fingertip pairs that no performance difference emerged.

\section{Discussion}\label{sec:discussion}

Across variations in horizontal approach distances, vertical finger displacements, and index-finger rotation, the mixed pair showed higher aggregate success than the uniform pair. Bootstrap confidence intervals for $\Delta S$ excluded zero in Experiments~1-1, 1-2, 1-3, and~2, and the CMH and threshold-sweep analyses supported the same direction in Experiments~1-1 and~2. By contrast, both fingertip pairs achieved perfect success after parameters were tuned for each coin size in Experiment~3. The mixed-stiffness design therefore broadened tolerance to selected parameter variations rather than increasing peak performance under favorable settings.

\begin{figure}[!t]
    \centering
    \includegraphics[width=0.5\linewidth]{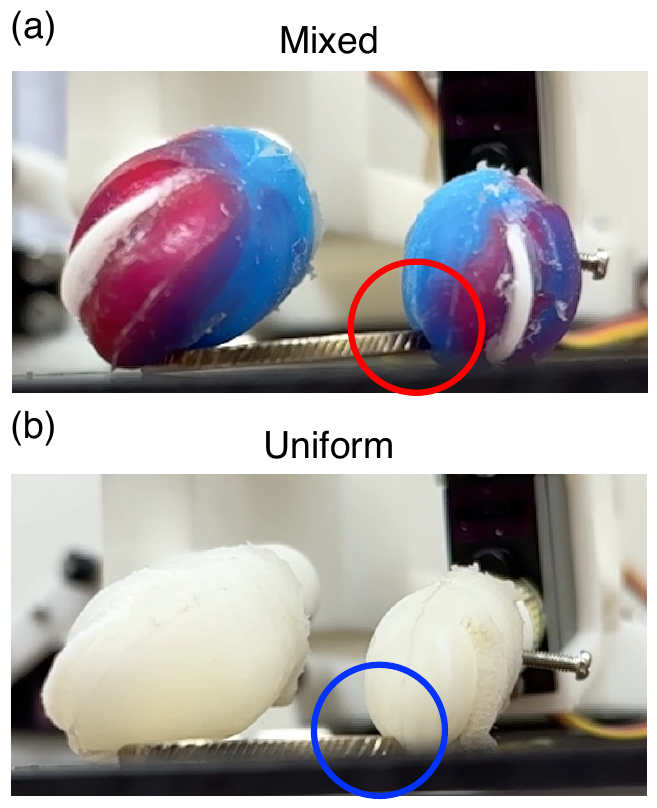}
    \caption{Representative fingertip deformation at $d_1 = 15$\,mm, $d_2 = 9$\,mm in Experiment~1-1. Success rates under this condition were 1.0 for the mixed pair and 0.2 for the uniform pair. Images were captured at the end of stage~3 (after index-finger rotation) and are intended for qualitative comparison of deformation patterns.}
    \label{fig:geometric}
\end{figure}

\paragraph{Lateral deformation-confinement hypothesis.}
With uniform stiffness, the pad may deform freely in the lateral direction when pressed against the coin edge, bulging toward the coin's upper surface and enveloping less of its lateral surface. In the mixed pair, the stiff A60 lateral regions may suppress such bulging and redirect deformation around the coin edge, and this confinement may stabilize the geometric constraint. In one condition, where the mixed pair succeeded in 10/10 trials and the uniform pair in only 2/10, deformation along the coin's lateral surface was visually more apparent for the mixed pair (Fig.~\ref{fig:geometric}). Outcomes in Experiment~1-1 tracked the number of soft--hard boundaries present at the contact: the nail-free pair had neither and failed in all 36~conditions, the uniform pair had the nail--pad boundary alone and succeeded at least once in 21~conditions, and the mixed pair had both the nail--pad and lateral boundaries and succeeded in 30. The lateral boundary may thus provide a further site at which pad deformation is arrested, functionally analogous to but mechanically distinct from the nail--pad boundary of our previous work\cite{A_Kumagai_23}: the nail is a quasi-rigid element supporting the whole fingertip dorsally, whereas the A60 regions are a stiff but still deformable elastomer that acts locally and laterally. Because the nail-free pair failed everywhere, we interpret the lateral boundary as complementing the nail--pad constraint rather than replacing it. The three fingertips also differed in effective overall stiffness, in lateral compliance, and in contact geometry, and neither contact forces nor deformation fields were measured. This ordering therefore does not isolate the boundaries as the cause, and how the position, orientation, and number of such boundaries affect constraint formation remains for future work.

\paragraph{Stage dependence and trade-off.}
Separating the process into two stages revealed many settings in which a geometric constraint formed but the transition to a stable grasp failed. The proportional reduction in high-success conditions was smaller for the mixed pair than for the uniform pair (60.0\% vs.\ 76.5\%). Together with the direct comparison of the 14 common conditions, this suggests that the mixed-stiffness advantage extended beyond initial constraint formation to constraint maintenance during transition. Failures often occurred when the coin slipped beneath the thumb, possibly because the fixed $65^\circ$ thumb-axis angle hindered the formation of a thumb-side constraint.

The mixed distribution was not uniformly advantageous. When coin position $l$ varied, the uniform pair showed the larger integrated success score. This parameter dependence is itself informative: a purely global stiffness effect would be expected to produce a more uniform trend across parameters, whereas an effect associated with a fixed material boundary could depend more directly on where contact occurs relative to that boundary. The observed pattern (an advantage under $d_1$, $d_2$, $r_1$, $r_2$, and $\phi$ but not under $l$) is therefore consistent with a spatially localized rather than a global stiffness effect, although contact geometry also changes with $l$ and the two cannot be separated here. The A60 regions may also reduce conformity as contact shifts along the finger, and the position of the material boundary and specimen-specific manufacturing variation may contribute as well. Thus, the design appears most beneficial for parameter variations that strongly alter pad deformation and local contact geometry.

\paragraph{Limitations and future work.}\label{sec:limitations}
This study evaluated only coins, and only with a single pinch strategy. Only one independently fabricated finger pair was tested for each condition, so between-specimen reproducibility remains unknown. The apparent modulus ratio of approximately 17 measured previously for uniform fingertips\cite{A_Kumagai_23} exceeds the several-fold regional differences reported in human fingertips\cite{A_Perez-Gonzalez_13} and does not directly represent the local ratio in the mixed fingertip. Moreover, adding A60 regions changes both the spatial distribution and the overall effective stiffness. Isolating the effect of patterning itself will require uniformly stiff or equivalent-overall-stiffness controls, as well as multiple independently fabricated specimens. In addition, because the coin initially lay flush against the plate, lifting was necessarily initiated without inserting any element beneath it; whether the compliant skin subsequently entered the small gap opened by the rotation could not be resolved from the recorded video and remains unverified. Future work should quantify contact pressure and surface deformation using pressure-sensitive films and digital image correlation, optimize stiffness ratios and boundary layouts, and test cards and other thin objects under additional pinch strategies.

\section{Conclusion}\label{sec:conclusion}

Inspired by the high-stiffness lateral sides and low-stiffness central pad of the human fingertip, we fabricated a mixed-stiffness anthropomimetic fingertip with E10 silicone at the center and A60 silicone at the sides, and compared its coin-grasping performance against a fingertip with uniform E10 skin.
The mixed pair maintained higher success rates across a broader range of explored conditions than the uniform pair under variations in horizontal approach distances, vertical finger displacements, and index-finger rotation angle.
This trend was observed not only for the operational criterion of geometric-constraint formation but also for the transition to a stable grasp.
Under tuned conditions, however, both fingertip pairs successfully grasped all six denominations of Japanese coins, and the uniform pair performed better when coin position was varied.
Thus, the mixed-stiffness design did not improve peak performance under tuned conditions, but increased tolerance to specific variations in the operating parameters.
Because the nail-free pair failed under all tested conditions, the nail remained essential in the present configuration, and we interpret the lateral stiff regions as complementing rather than replacing the nail--pad constraint of our previous work.
More broadly, adding a passive material stiffness boundary within a nail-supported pad can broaden the range of conditions under which a geometric constraint forms, without requiring active stiffness control, making boundary placement a candidate fingertip design variable.

\section*{Author Contributions}
Kaigen Go: Conceptualization, Methodology, Investigation, Formal analysis, Data curation, Visualization, and Writing---original draft. Yinlai Jiang: Supervision, Validation, and Writing---review and editing. Hiroshi Yokoi: Supervision, Validation, and Writing---review and editing. Shunta Togo: Conceptualization, Methodology, Formal analysis, Funding acquisition, Project administration, Supervision, and Writing---review and editing. All authors reviewed and approved the final manuscript.

\section*{Statements and Declarations}

\subsection*{Ethical Considerations}
Ethical approval was not required because this study did not involve human participants, human data, human tissue, or animals.

\subsection*{Consent to Participate}
Not applicable.

\subsection*{Consent for Publication}
Not applicable.

\subsection*{Declaration of Conflicting Interest}
The authors declared no potential conflicts of interest with respect to the research, authorship, and/or publication of this article.

\subsection*{Funding Statement}
This work was supported in part by JSPS KAKENHI Grant Numbers JP26K00910 and JP23H00166.

\subsection*{Data Availability}
The materials supporting this study are publicly available in the GitHub repository (\url{https://github.com/TogoLab/mixed-stiffness-fingertip-coin-grasping}). These materials include Video~S1, the source code and data used for the statistical analyses, the Arduino control source code, 3D CAD files, and 3D-printable part files. The condition-wise trial outcomes used to generate the heat maps are included with the statistical-analysis source code; therefore, no separate Data~S1 file is provided.


\clearpage
\setcounter{figure}{0}
\setcounter{table}{0}
\renewcommand{\thefigure}{S\arabic{figure}}
\renewcommand{\thetable}{S\arabic{table}}

\section*{Supplementary Information}

\noindent\textbf{Video S1.} Representative footage of coin grasping with the mixed-stiffness fingertip pair, including geometric-constraint formation and transition to a stable grasp (available in the GitHub repository).

\noindent\textbf{Code S1.} Source code and condition-wise trial data used for the statistical analyses and generation of the heat maps (available in the GitHub repository).

\noindent\textbf{Code S2.} Arduino source code used to control the automated grasping apparatus (available in the GitHub repository).

\begin{figure}[!htbp]
    \centering
    \includegraphics[width=0.95\linewidth]{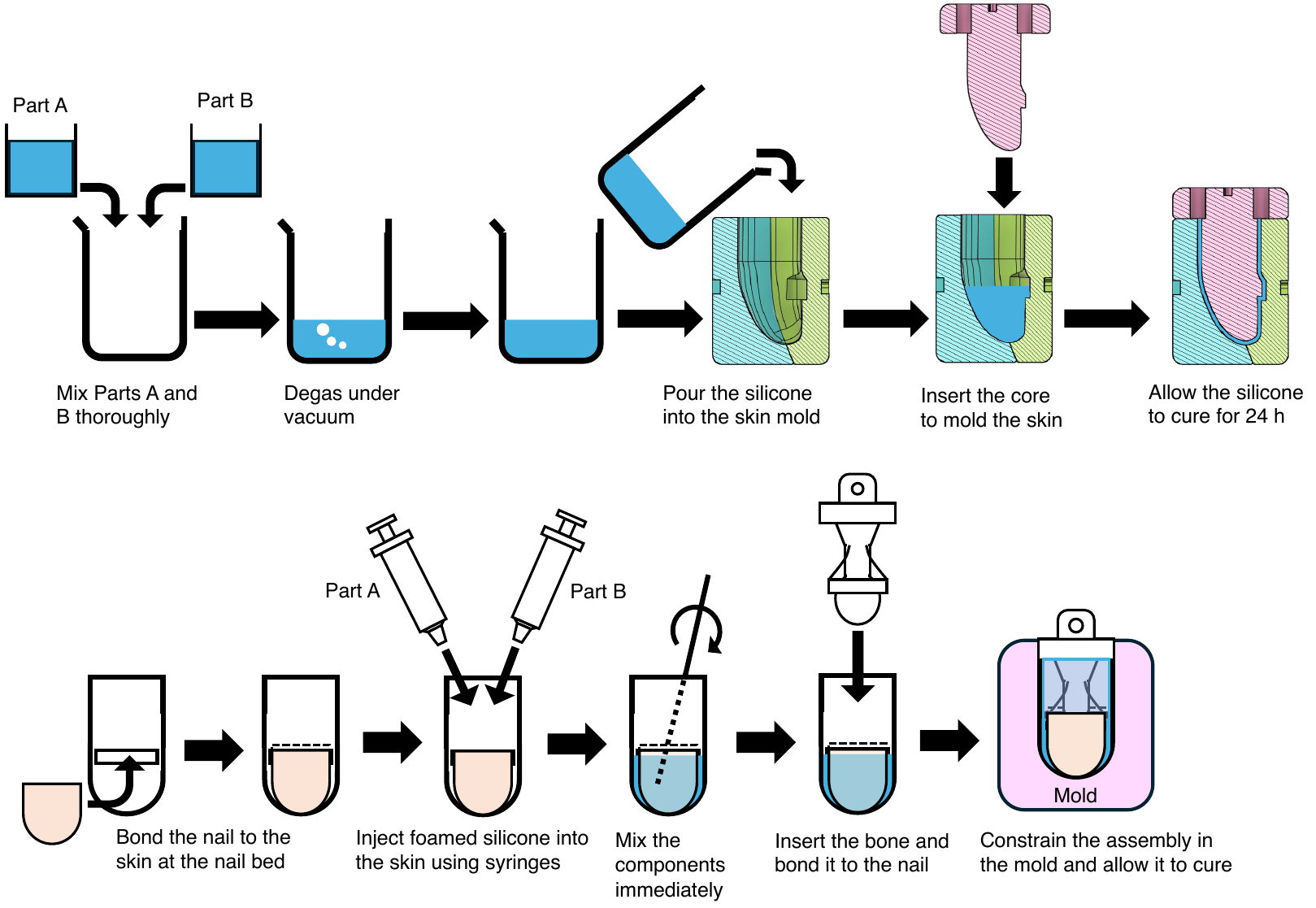}
    \caption{
    Fabrication procedure for the uniform-stiffness fingertip.
    Parts A and B of the Shore~E10 silicone were mixed 1:1, injected into the gap between the outer mold
    and the inner core, and cured for 24~h to form a homogeneous skin.
    After curing, the 3D-printed nail was bonded to the skin.
    Foamed silicone, mixed at a 100:47 A:B weight ratio, was then injected into the skin as subcutaneous tissue,
    and the 3D-printed bone was quickly inserted to complete the artificial fingertip before the foam set.
    The assembly was constrained in the skin-forming mold during curing
    to reduce dimensional distortion caused by silicone foaming.
    }
    \label{fig:supp_uniform_fabrication}
\end{figure}

\clearpage

\begin{figure}[!htbp]
    \centering
    \includegraphics[width=0.95\linewidth]{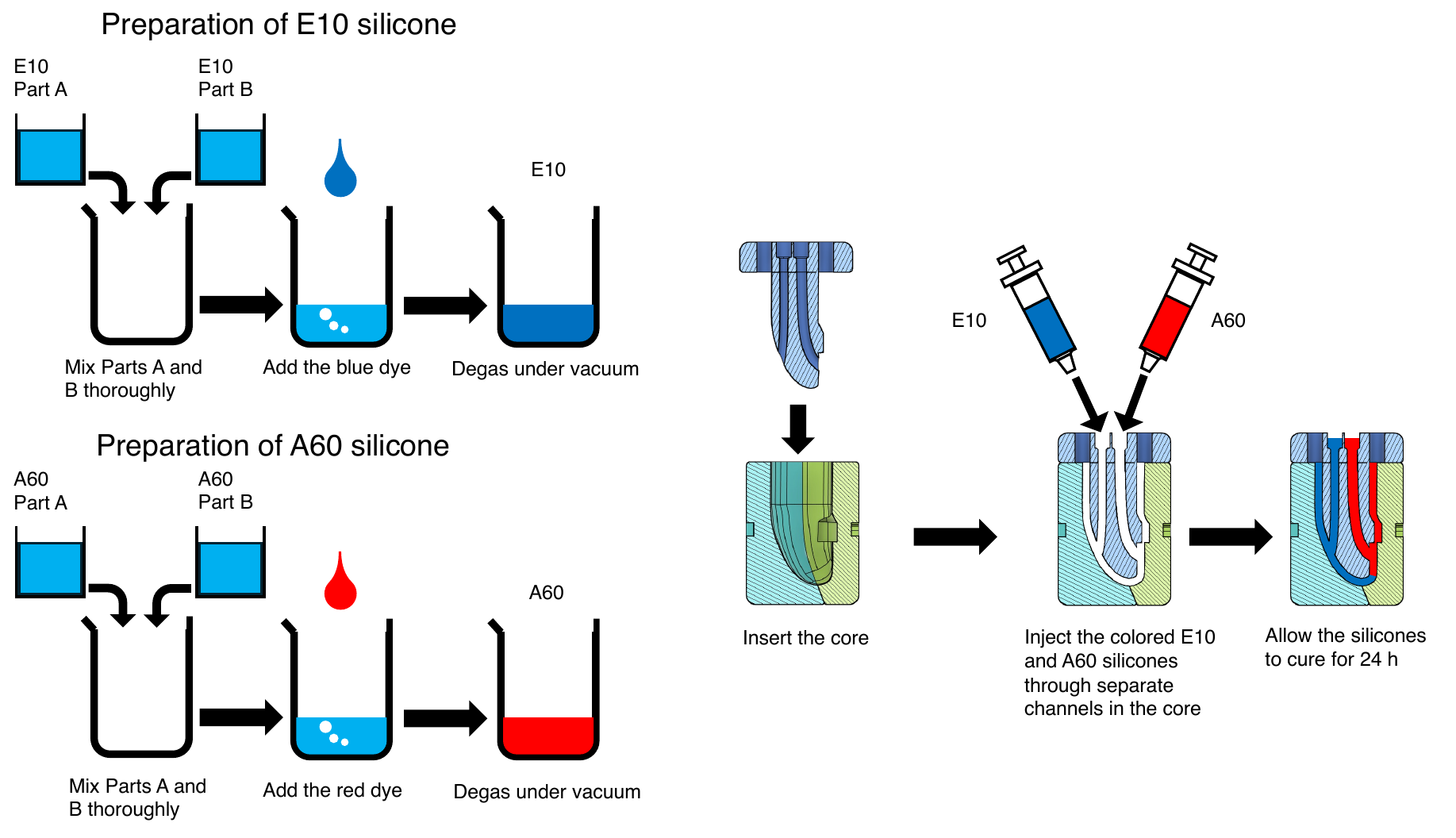}
    \caption{
    Fabrication procedure for the mixed-stiffness fingertip.
    Two independent injection channels were incorporated into the inner core.
    Shore~E10 silicone was injected into the central pad region,
    whereas Shore~A60 silicone was injected into the lateral regions.
    Parts A and B of each skin silicone were mixed 1:1. The two materials were injected separately and cured simultaneously for 24~h
    to form a continuous skin with a soft center and stiff sides.
    The nail, foamed-silicone subcutaneous tissue, and bone were subsequently
    assembled using the same procedure as for the uniform fingertip.
    }
    \label{fig:supp_mixed_fabrication}
\end{figure}

\clearpage

\begin{figure}[!htbp]
    \centering
    \includegraphics[width=\linewidth]{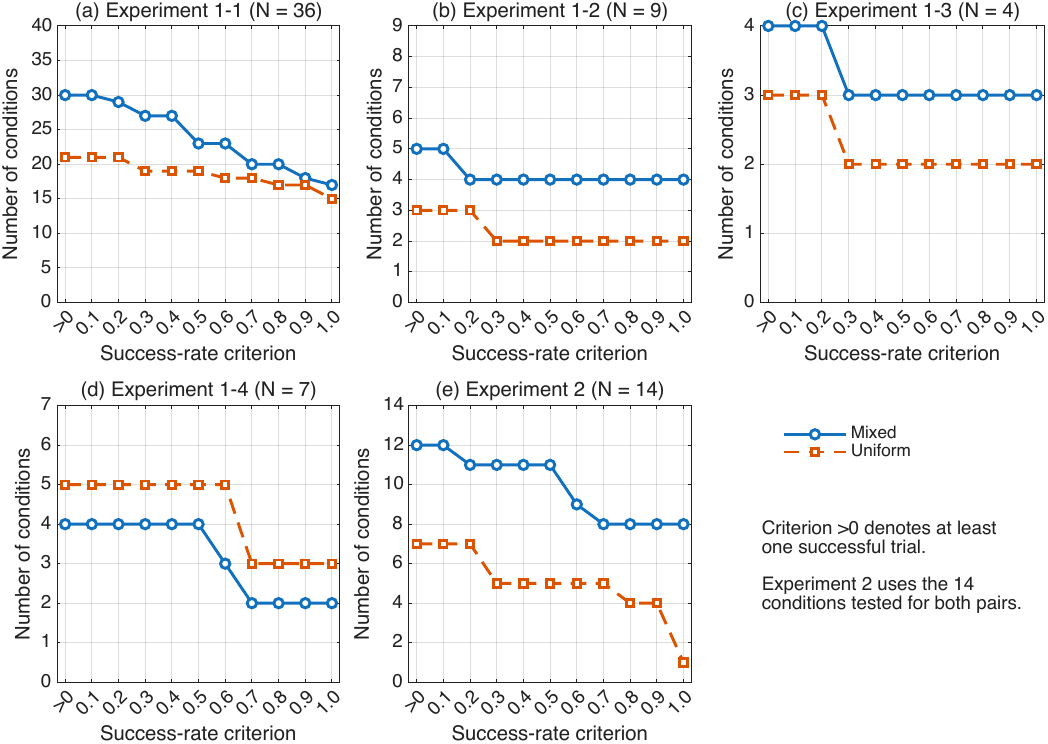}
    \caption{
    Threshold-sweep analysis of the successful operating range.
    Each panel shows the number of parameter conditions meeting each
    success-rate criterion for the Mixed and Uniform fingertip pairs:
    (a)~Experiment~1-1,
    (b)~Experiment~1-2,
    (c)~Experiment~1-3,
    (d)~Experiment~1-4, and
    (e)~Experiment~2.
    The first criterion ($>0$) counts conditions with at least one
    successful trial; the subsequent criteria count conditions with
    success rates greater than or equal to
    $0.1, 0.2, \ldots, 1.0$.
    Experiment~2 includes only the 14 parameter conditions tested
    with both fingertip pairs.
    }
    \label{fig:supp_threshold_sweep}
\end{figure}

\begin{table}[!htbp]
    \centering
    \caption{
    Sensitivity of the Mixed--Uniform difference to weighting for the
    nonuniform $d_1$--$d_2$ sampling grid.
    Grid weights were calculated from midpoint-based cell widths along
    $d_1$ and $d_2$ and normalized over the parameter conditions available
    for both fingertip pairs.
    Here, $\Delta$ denotes the Mixed value minus the Uniform value.
    The direction of the difference was unchanged after grid weighting.
    }
    \label{tab:supp_weighted_grid_sensitivity}
    \small
    \setlength{\tabcolsep}{5pt}
    \begin{tabular}{@{}lrrrrrr@{}}
        \toprule
        &
        \multicolumn{3}{c}{Unweighted mean success rate} &
        \multicolumn{3}{c}{Grid-weighted mean success rate}
        \\
        \cmidrule(lr){2-4}
        \cmidrule(lr){5-7}
        Experiment
        & Mixed
        & Uniform
        & $\Delta$
        & Mixed
        & Uniform
        & $\Delta$
        \\
        \midrule
        Experiment~1-1
        & 0.653
        & 0.514
        & 0.139
        & 0.632
        & 0.510
        & 0.122
        \\
        Experiment~2
        & 0.693
        & 0.350
        & 0.343
        & 0.692
        & 0.325
        & 0.367
        \\
        \bottomrule
    \end{tabular}
\end{table}

\clearpage

\end{document}